\documentclass[letterpaper]{article}
\usepackage{cuted}
\usepackage[preprint]{aaai2027}
\usepackage[hyphens]{url}
\usepackage{graphicx}
\usepackage{natbib}
\usepackage{caption}
\usepackage{amsmath}
\usepackage{array}
\usepackage{booktabs}
\usepackage{calc}
\usepackage{algorithm}
\usepackage{algpseudocode}
\usepackage{multirow}

\DeclareCaptionStyle{ruled}{labelfont=normalfont,labelsep=colon,strut=off}
\newcommand\Description[2][]{}
\newcommand{\AppendixColumnBreak}{\newpage}

\title{XSPA: Crafting Imperceptible X-Shaped Sparse Adversarial Perturbations for Transferable Attacks on VLMs}
\author{Chengyin Hu\textsuperscript{1}, Jiaju Han\textsuperscript{1,2}, Xuemeng Sun\textsuperscript{1}, Qike Zhang\textsuperscript{1}, Luwei Yang\textsuperscript{2},\\
Lehan Sun\textsuperscript{1}, Jiahuan Long\textsuperscript{3}, Yiwei Wei\textsuperscript{4}, Jiujiang Guo\textsuperscript{4,5}}
\affiliations{\textsuperscript{1}China University of Petroleum-Beijing at Karamay\\
\textsuperscript{2}Shenzhen Research Institute of Big Data \quad
\textsuperscript{3}Shanghai Jiao Tong University\\
\textsuperscript{4}Tianjin University \quad
\textsuperscript{5}North University of China}

\begin{document}

\maketitle

\begin{abstract}
Vision-language models (VLMs) rely on a shared visual-textual representation space to support tasks such as zero-shot classification, image captioning, and visual question answering (VQA). Although this shared space enables strong cross-task generalization, it also creates a critical vulnerability: subtle visual perturbations may propagate through the common embedding space and induce correlated semantic failures across tasks. This issue is especially concerning in interactive and decision-support scenarios, yet it remains unclear whether VLMs are still fragile under highly constrained, sparse, and geometrically fixed perturbations. We propose X-shaped Sparse Pixel Attack (XSPA), an imperceptible structured attack that restricts perturbations to two intersecting diagonal lines. Compared with dense perturbations and flexible localized patches, XSPA operates under a much stricter attack budget, offering a stringent test of VLM robustness. Within this sparse support, XSPA jointly optimizes a classification-oriented objective, cross-task semantic guidance, and regularization on perturbation magnitude and linewise smoothness, inducing both transferable misclassification and semantic drift in captioning and VQA while preserving visual subtlety. Under the default setting, XSPA modifies only about 1.04\% of image pixels. Experiments on the COCO dataset show that XSPA consistently degrades performance across zero-shot classification, image captioning, and VQA. Zero-shot accuracy drops by 52.33 points on OpenAI CLIP ViT-L/14 and 67.00 points on OpenCLIP ViT-B/16, while GPT-4-evaluated caption consistency decreases by up to 58.60 points and VQA correctness by up to 44.25 points. These results show that even highly sparse and visually subtle perturbations with fixed geometric priors can substantially disrupt cross-task semantics in VLMs, revealing an important robustness gap in current multimodal systems.
\end{abstract}

\begin{figure*}[t]
  \centering
  \includegraphics[width=\textwidth]{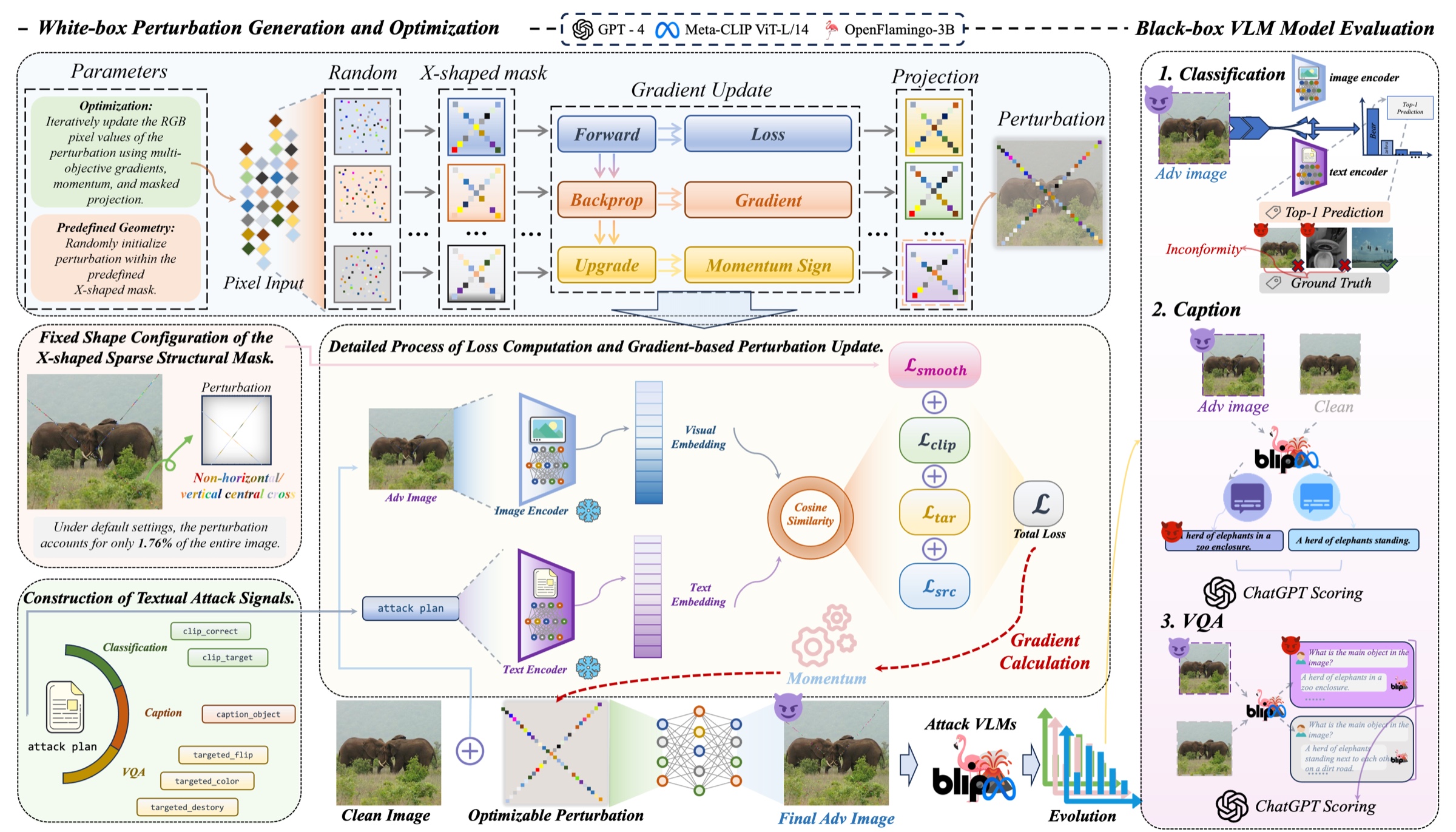}
  \Description{Overview of the XSPA pipeline. A sparse X-shaped perturbation is optimized on a surrogate CLIP model and then transferred to zero-shot classification, image captioning, and visual question answering models.}
  \caption{Overall framework of XSPA. XSPA optimizes perturbations within a fixed X-shaped sparse mask on a surrogate CLIP model, integrating momentum-based optimization, input diversity, and joint semantic objectives to promote target-oriented semantic attraction while suppressing source semantics. The resulting adversarial image is then transferred to downstream vision-language models for evaluation across zero-shot classification, image captioning, and visual question answering tasks.}
  \label{fig:framework}
\end{figure*}

\section{Introduction}
Recent advances in multimodal learning, driven by models such as CLIP and BLIP-2, have substantially improved cross-modal understanding in zero-shot classification, visual question answering, and open-ended generation \cite{radford2021_learning,li2023_blip}. A key reason is the shared embedding space aligning visual and textual representations across downstream tasks. Yet this same mechanism also introduces a robustness concern. Perturbations affecting visual encoding may propagate through the shared semantic space and cause not only isolated prediction errors, but also cross-task semantic failures across classification, captioning, and question answering. This matters because a single corrupted input can affect multiple decisions at once.

Adversarial attack research has consistently shown that deep models are highly vulnerable to perturbations that are nearly imperceptible to humans \cite{goodfellow2015_explaining,madry2018_deep}. Existing attacks on visual models mainly fall into two categories: dense full-image perturbations under $L_p$ constraints, which prioritize attack strength, and localized or physically motivated interventions, such as patches, shadows, or optical manipulations, which emphasize locality and real-world realizability \cite{brown2018_adversarial,zhong2022_shadows,hsiao2024_natural}. Although these studies provide valuable insights, they usually retain substantial freedom in perturbation support, shape, or optimization space. Such settings demonstrate attack effectiveness, but are less informative for a more fundamental question: how vulnerable are VLMs when perturbation support, geometry, and optimization freedom are all restricted?

This question is particularly important for vision-language models. Unlike unimodal classifiers, VLMs must not only produce correct visual predictions, but also preserve semantic alignment across multiple downstream tasks. As a result, robustness evaluation under extremely flexible perturbation settings may obscure whether cross-task semantic behavior can already be destabilized by visual signals that are both extremely low-budget and strongly structured. Characterizing this vulnerability boundary is valuable in two respects: scientifically, it clarifies how much geometric freedom is required to disrupt multimodal alignment; practically, it provides a controlled stress test for future robustness evaluation and defense design.

Motivated by this perspective, we study whether a perturbation with extremely limited support and fixed geometry is already sufficient to induce significant cross-task semantic disruption in VLMs. Rather than treating perturbation shape as a free variable, we impose a stringently constrained geometric prior and use it as a controlled probe of multimodal robustness. Concretely, we propose X-shaped Sparse Pixel Attack (XSPA), which confines perturbations to a sparse X-shaped structure formed by two intersecting diagonal line segments. Within this restricted region, XSPA integrates transfer-enhancing optimization strategies \cite{author2019_improving} with joint semantic objectives for zero-shot classification, image captioning, and visual question answering. Beyond inducing label errors, the optimization promotes target-oriented semantic attraction and suppresses source semantics, thereby encouraging semantic drift across downstream generative tasks. Figure~\ref{fig:framework} illustrates this pipeline, from sparse perturbation optimization on the surrogate CLIP model to transfer-based evaluation on downstream vision-language tasks.

We evaluate XSPA on four CLIP-style visual encoders and six downstream VLMs, using LLM-as-a-Judge style automatic evaluators for open-ended tasks \cite{zheng2023_judging,openai2024_gpt4o}. The results show that XSPA markedly degrades performance on zero-shot classification, image captioning, and visual question answering under a severely restricted perturbation budget. These findings suggest that even when perturbations are confined to an extremely small region with a fixed geometric prior, VLMs remain vulnerable to cross-task semantic disruption. More broadly, our results show that multimodal robustness cannot be understood only through dense noise or flexible local attacks; it must also account for the fragility of shared semantic alignment under strongly constrained structured perturbations. In this sense, XSPA serves not only as an effective attack, but also as a controlled diagnostic tool for probing how little geometric freedom is needed to destabilize multimodal behavior. This perspective helps connect attack evaluation with a more precise characterization of vulnerability boundaries in shared visual-textual representation spaces. The main contributions of this work are as follows.
\begin{itemize}
\item To the best of our knowledge, we propose the first X-shaped sparse pixel attack (XSPA) that constrains perturbations to a fixed geometric structure with extremely limited support. The method unifies momentum-based optimization, input diversity, cross-task semantic guidance, and source-semantic suppression within a highly restricted two-diagonal-line perturbation region.
\item We conduct extensive experiments across zero-shot classification, image captioning, and visual question answering tasks, demonstrating that XSPA consistently achieves substantial cross-task semantic disruption on four CLIP-style encoders and six downstream vision-language models under minimal perturbation budgets.
\item We perform comprehensive ablation studies that reveal how highly constrained structured perturbations with fixed geometry can still induce strong multimodal misalignment, providing insights into the vulnerability boundaries of shared visual-textual representation spaces.
\end{itemize}

\section{Related Work}
\label{sec:related}

\subsection{Sparse and Structured Attacks on Vision Models}

Adversarial attacks on vision models provide the methodological basis for studying attacks on vision-language models. Early work established that deep networks remain vulnerable under diverse threat models \cite{goodfellow2015_explaining,madry2018_deep,author2016_deepfool,author2019_improving,andriushchenko2020_square}. Most relevant to our work are sparse and structured attacks that modify only a few pixels or constrained regions \cite{su2019_one,modas2019_sparsefool,zhong2025_sparse}, and physically inspired attacks using patches, shadows, or natural illumination \cite{brown2018_adversarial,ran2023_cross,zhong2022_shadows,huang2022_spaa,hsiao2024_natural}. However, most existing methods still optimize support size, patch area, or physical realizability rather than testing highly constrained geometric priors; by contrast, we fix an X-shaped sparse structure and test whether it alone suffices to induce cross-task semantic deviation.

\subsection{Attacks and Robustness Studies on Vision-Language Models}

With the rise of CLIP-style vision-language pre-training, robustness in shared cross-modal embedding spaces has become a central issue \cite{radford2021_learning}. Prior work shows that adversarial perturbations can transfer across retrieval, classification, and generation through the shared representation space, and that large VLMs remain vulnerable in both classification-oriented and open-ended settings \cite{zhang2022_adversarial,author2023_set,zhao2023_evaluating,cui2024_robustness}. Recent work has improved attack transferability and unified cross-task design \cite{zhang2025_anyattack,xie2025_chain,author2025_gleam,zhao2025_one,rusia2025_when,li2026_transform,li2026_mpcattack}, while robustness studies explore certification, adaptation, and test-time tuning \cite{wang2024_mmcert,schlarmann2024_robust,tong2025_zero,sheng2025_tpt,author2025_improving,mirza2026_energy}.

\subsection{Automatic Evaluation for Open-Ended Tasks}

For zero-shot classification, adversarial effectiveness can be measured directly by accuracy or attack success rate, but for open-ended tasks such as image captioning and visual question answering, exact-match and lexical-overlap metrics often fail to capture semantically meaningful changes \cite{zhao2023_evaluating,cui2024_robustness}. We therefore adopt a GPT-4-based automatic evaluator that assesses semantic correctness and factual consistency, providing a more faithful measurement of whether perturbations alter the model's understanding rather than merely changing surface wording. This distinction is essential in cross-task attack evaluation: a perturbation may preserve surface fluency while changing the predicted concept, answer, or factual relation. Combining task-specific metrics with semantic judging therefore allows classification, captioning, and VQA outcomes to be compared under a unified evaluation protocol \cite{zheng2023_judging,openai2024_gpt4o}.

\section{Method}
\label{sec:method}

\subsection{Problem Definition}

Given an input image \(x\in[0,1]^{3\times H\times W}\), XSPA optimizes a perturbation \(\delta\) restricted to a fixed X-shaped binary mask \(M\in\{0,1\}^{H\times W}\):
\begin{equation}
x^{adv}=\operatorname{clip}(x+M\odot\delta,0,1),
\end{equation}
where \(\odot\) denotes element-wise multiplication. We consider a white-box surrogate / black-box downstream setting: the perturbation is optimized on a CLIP-style surrogate and transferred to downstream models for zero-shot classification, image captioning, and visual question answering. Thus, unlike full-image \(L_p\)-bounded attacks, XSPA modifies only a highly sparse predefined support.

\subsection{X-shaped Sparse Structural Mask}

XSPA uses two oblique line segments crossing at \(c_{\mathrm{ctr}}=(\rho_{\mathrm{col}}W,\rho_{\mathrm{row}}H)\). Given their angles \(\mathcal{A}=\{\theta_1,\theta_2\}\), relative length \(r\), and width \(b\), the mask support is
\begin{equation}
\Omega_M=
\bigcup_{\theta\in\mathcal{A}}
\mathcal{N}_b\bigl(\ell(c_{\mathrm{ctr}},\theta,r)\bigr),
\end{equation}
where \(\ell(\cdot)\) is a line segment whose length is proportional to the shorter image side and \(\mathcal{N}_b(\cdot)\) expands it to width \(b\). The binary mask and its sparsity ratio are
\begin{equation}
M_{h,w}=\mathbf{1}[(h,w)\in\Omega_M],
\qquad
\eta_M=\frac{|\Omega_M|}{HW}.
\end{equation}
The mask is obtained by rasterizing the two centerlines and expanding their sampled coordinates locally. Under the default \(384\times384\) setting, the 1,532-pixel support gives \(\eta_M\approx1.04\%\). Further construction details are provided in the Appendix.

Fixing the support before optimization separates geometric effects from unconstrained pixel selection and prevents the optimizer from allocating perturbations to visually convenient regions. Consequently, transferability must arise from the learned values within the prescribed X structure rather than from adaptive support search.

\subsection{Cross-task Joint Objective}

On the fixed support, XSPA jointly optimizes classification attack signals, cross-task semantic guidance, and structural regularization:
\begin{equation}
\mathcal{L}
=
\lambda_{clip}^{(t)}\mathcal{L}_{clip}
+
\lambda_{tar}^{(t)}\mathcal{L}_{tar}
+
\lambda_{src}^{(t)}\mathcal{L}_{src}
+
\mathcal{L}_{smooth}.
\end{equation}
Let \(f_v(\cdot)\) and \(f_t(\cdot)\) be the surrogate CLIP image and text encoders, \(v=f_v(x^{adv})\) the normalized image feature, and \(z_k=\langle v,f_t(s_k)\rangle/\tau\) the logit for class prompt \(s_k\). We use a margin loss for untargeted attacks and cross-entropy for targeted attacks:
\begin{equation}
\mathcal{L}_{clip}=
\begin{cases}
z_y-\max_{k\neq y}z_k, & \text{untargeted},\\
\mathrm{CE}(z,y_t), & \text{targeted},
\end{cases}
\end{equation}
where \(y\) and \(y_t\) denote the source and target classes.

To extend the attack beyond label prediction, we construct a target text pool \(\mathcal{T}_{tar}=\{p_u\}_{u=1}^{N_t}\) and a source text pool \(\mathcal{T}_{src}=\{q_w\}_{w=1}^{N_s}\) from target labels, caption-side semantic drifts, and VQA-side answer shifts. Their losses are
\begin{equation}
\begin{aligned}
\mathcal{L}_{tar}
&=-\frac{1}{N_t}\sum_{u=1}^{N_t}\langle v,f_t(p_u)\rangle,\\
\mathcal{L}_{src}
&=\frac{1}{N_s}\sum_{w=1}^{N_s}\langle v,f_t(q_w)\rangle.
\end{aligned}
\end{equation}
The first term attracts the adversarial representation toward target semantics, whereas the second suppresses source semantics. Combining both pools avoids relying on a single class label and exposes the surrogate representation to task-relevant concepts shared by classification, captioning, and VQA. Because all prompts are encoded by the same frozen text encoder, these signals remain comparable in one embedding space and define a task-agnostic semantic direction that can transfer beyond the surrogate classifier. To avoid sharp local oscillations on the thin support, we further use
\begin{equation}
\mathcal{L}_{smooth}
=
\lambda_{mag}^{(t)}\mathcal{L}_{mag}
+
\lambda_{line}^{(t)}\mathcal{L}_{line},
\end{equation}
where
\begin{equation}
\begin{aligned}
\mathcal{L}_{mag}
&=\frac{1}{|\Omega_M|}
\sum_{(h,w)\in\Omega_M}\|\delta_{h,w}\|_2^2,\\
\mathcal{L}_{line}
&=\frac{1}{|\mathcal{P}|}
\sum_{P\in\mathcal{P}}\frac{1}{|P|-1}
\sum_{n=2}^{|P|}
\|\delta_{P_n}-\delta_{P_{n-1}}\|_2^2.
\end{aligned}
\end{equation}
The magnitude term controls isolated perturbation spikes, whereas the line term couples neighboring pixels along each ordered centerline path \(P\in\mathcal{P}\). Their combination preserves the sparse geometry while discouraging visually abrupt changes. For each \(\nu\in\{clip,tar,src,mag,line\}\), the corresponding weight follows
\begin{equation}
\lambda_{\nu}^{(t)}
=
\begin{cases}
\lambda_{\nu}^{A}, & t<\gamma N,\\
\lambda_{\nu}^{B}, & t\ge\gamma N.
\end{cases}
\end{equation}
Early iterations emphasize classification guidance to stabilize the attack direction, whereas later iterations strengthen semantic misalignment to promote cross-task transfer.

\subsection{Mask-constrained Optimization}

The perturbation is restricted to the feasible set
\begin{equation}
\mathcal{C}(M,\epsilon)
=
\left\{\delta\;\middle|\;(1-M)\odot\delta=0,\ 
\|\delta\|_{\infty}\le\epsilon\right\}.
\end{equation}
At iteration \(t\), we apply random resize-and-pad input diversity \(D(\cdot)\), compute the joint gradient, accumulate MI-FGSM-style momentum, and project the update onto \(\mathcal{C}(M,\epsilon)\):
\begin{equation}
\begin{aligned}
g^{(t)}
&=\nabla_{\delta}\mathcal{L}\!\left(D(x^{adv,(t)})\right),\\
m^{(t+1)}
&=\mu m^{(t)}
+\frac{g^{(t)}}{\operatorname{mean}(|g^{(t)}|)+10^{-12}},\\
\delta^{(t+1)}
&=M\odot\operatorname{clip}\!\left(
\delta^{(t)}-\alpha\,\operatorname{sign}(m^{(t+1)}),
-\epsilon,\epsilon\right).
\end{aligned}
\end{equation}
Input diversity reduces surrogate overfitting, while momentum stabilizes the update direction. The projection keeps every modification inside the X-shaped support. The complete procedure and pseudocode are included in the Appendix.

\begin{table*}[h]
\centering
\caption{Zero-shot classification results under different adversarial attacks. Lower ACC and higher naturalness are better. Numbers in parentheses denote the absolute change relative to clean samples. Bold indicates the best result.}
\label{tab:zero-shot}
\small
\resizebox{0.94\textwidth}{!}{%
\begin{tabular}{@{}lcccccccc@{}}
\toprule
\multicolumn{1}{c}{} & \multicolumn{2}{c}{\textbf{OpenCLIP ViT-B/16}} & \multicolumn{2}{c}{\textbf{Meta-CLIP ViT-L/14}} & \multicolumn{2}{c}{\textbf{EVA-CLIP ViT-G/14}} & \multicolumn{2}{c}{\textbf{OpenAI CLIP ViT-L/14}} \\
\cmidrule(lr){2-3} \cmidrule(lr){4-5} \cmidrule(lr){6-7} \cmidrule(lr){8-9}
\textbf{Method} & ACC(\%) & GPT-4o & ACC(\%) & GPT-4o & ACC(\%) & GPT-4o & ACC(\%) & GPT-4o \\
\midrule
Clean & 97 & 3.65 & 98 & 3.65 & 98 & 3.65 & 93 & 3.65 \\
NaturalLightAttack & 94($\downarrow$3) & 2.32($\downarrow$1.33) & 97($\downarrow$1) & 2.14($\downarrow$1.51) & 97($\downarrow$1) & 2.16($\downarrow$1.49) & 93(-) & 2.07($\downarrow$1.58) \\
ShadowAttack & 84($\downarrow$13) & 2.11($\downarrow$1.54) & 82($\downarrow$16) & 1.91($\downarrow$1.74) & 95($\downarrow$3) & 2.03($\downarrow$1.62) & 79($\downarrow$14) & 1.78($\downarrow$1.87) \\
ITA & 46($\downarrow$51) & 2.11($\downarrow$1.54) & 64($\downarrow$34) & 2.15($\downarrow$1.50) & 84($\downarrow$14) & 2.17($\downarrow$1.48) & 51($\downarrow$42) & 2.15($\downarrow$1.50) \\
\midrule
\textbf{Ours} & \textbf{30}($\downarrow$\textbf{67}) & \textbf{2.71}($\downarrow$\textbf{0.94}) & \textbf{60.67}($\downarrow$\textbf{37.33}) & \textbf{2.89}($\downarrow$\textbf{0.76}) & \textbf{63.71}($\downarrow$\textbf{34.29}) & \textbf{2.78}($\downarrow$\textbf{0.87}) & \textbf{40.67}($\downarrow$\textbf{52.33}) & \textbf{2.76}($\downarrow$\textbf{0.89}) \\
\bottomrule
\end{tabular}%
}
\end{table*}

\section{Experiments}
\label{sec:experiment}

\subsection{Experimental Setup}

\textbf{Dataset.} Following ITA's evaluation protocol, we select 300 images from COCO-80 \cite{lin2014_microsoft} as a unified evaluation set. All three tasks are evaluated on the same dataset to enable consistent comparison across attack methods. For zero-shot classification, we use the COCO-80 category set as the semantic label space.

\textbf{Models.} For zero-shot classification, we evaluate four representative CLIP-style visual encoders: OpenCLIP ViT-B/16 \cite{cherti2023_openclip}, Meta-CLIP ViT-L/14 \cite{xu2024_metaclip}, EVA-CLIP ViT-G/14 \cite{sun2023_eva}, and OpenAI CLIP ViT-L/14 \cite{radford2021_learning}. For downstream multimodal tasks, we further evaluate six mainstream vision-language models, including LLaVA-1.5, LLaVA-1.6 \cite{liu2023_visual}, OpenFlamingo \cite{awadalla2023_openflamingo}, BLIP-2 (FlanT5XL ViT-L), BLIP-2 (FlanT5XL) \cite{li2023_blip}, and InstructBLIP (FlanT5XL) \cite{dai2023_instructblip}, in order to examine the transferability of XSPA across different model architectures and task settings.

\textbf{Baselines.} We compare XSPA with NaturalLightAttack, ShadowAttack, and ITA \cite{hsiao2024_natural,zhong2022_shadows,rusia2025_when}. These baselines cover natural-light, shadow-based, and illumination-transformation attacks. Their threat models do not perfectly match the fixed sparse geometry of XSPA, so the comparison should be interpreted as a robustness-oriented reference rather than a strict like-for-like ranking. The purpose is to test whether VLMs remain vulnerable even when the perturbation support is restricted to a much simpler and more constrained structure.

\textbf{Evaluation Metrics.} For zero-shot classification, we report Top-1 accuracy and its absolute drop from clean samples. T-ASR is the rate of predictions driven to a predefined target, while U-ASR is the rate of untargeted misclassification; higher values indicate stronger attacks. GPT-4o evaluates image naturalness on a 4-point scale covering visual naturalness, physical consistency, and adversarial plausibility. For captioning and VQA, GPT-4 evaluates consistency and correctness under a fixed LLM-as-a-Judge prompt and rubric across all methods and models \cite{zheng2023_judging,openai2023_gpt4,openai2024_gpt4o}.

\textbf{Implementation Details.} Main experiments use 200 iterations, while ablations follow the frozen protocol below. XSPA is optimized on a CLIP-style surrogate and transferred to downstream VLMs. All experiments use a single NVIDIA RTX 4090 24GB GPU; the remaining optimization and loss scheduling follow Section 3.

\begin{figure}[h]
\centering
\captionsetup{skip=3pt}
\includegraphics[width=\columnwidth]{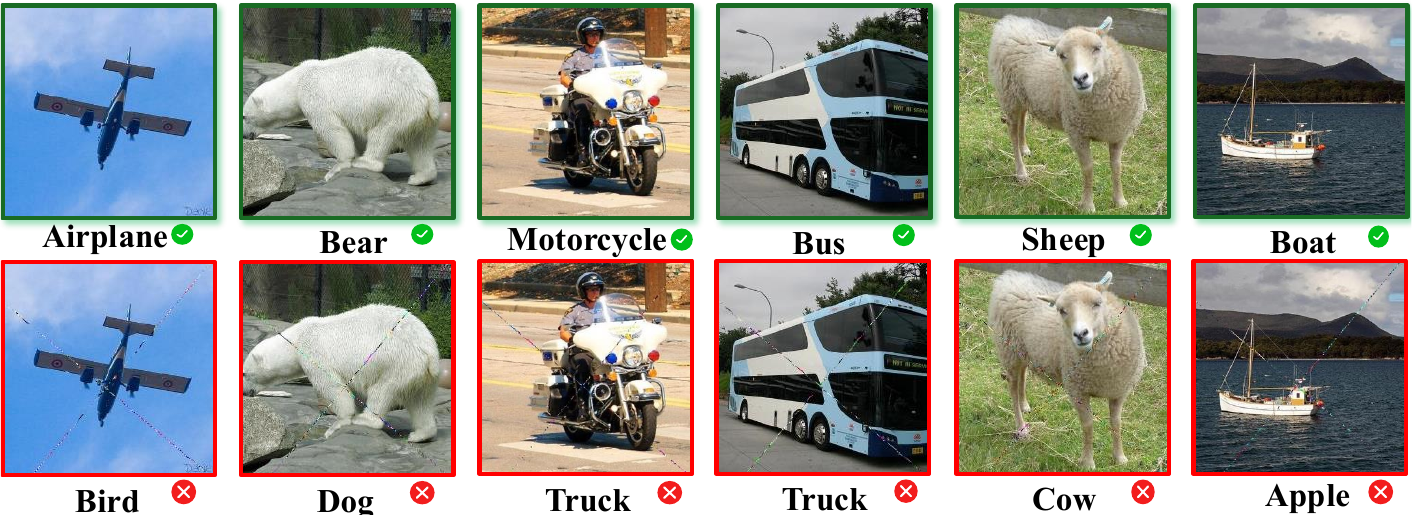}
\Description{Six qualitative zero-shot classification examples comparing clean and adversarial images. The adversarial row shows sparse X-shaped perturbations that preserve natural appearance while inducing wrong predictions.}
\caption{Qualitative zero-shot classification examples. The top row shows clean images, and the bottom row shows XSPA adversarial images. Despite the fixed X-shaped sparse perturbation, clear prediction shifts are observed.}
\label{fig:zero-shot-qual}
\end{figure}

\subsection{Main Results}

\noindent\textbf{Zero-Shot Classification.} Table~\ref{tab:zero-shot} summarizes the zero-shot classification results. Under this protocol, XSPA causes the largest accuracy drop on all four CLIP-style visual encoders while also achieving the highest GPT-4o naturalness scores. Specifically, it reduces the Top-1 accuracy to 30.00\%, 60.67\%, 63.71\%, and 40.67\% on OpenCLIP ViT-B/16, Meta-CLIP ViT-L/14, EVA-CLIP ViT-G/14, and OpenAI CLIP ViT-L/14, respectively. Relative to ITA, XSPA further decreases accuracy by 16.00, 3.33, 20.29, and 10.33 percentage points. These results show that even under a highly constrained X-shaped sparse support, CLIP image-text alignment can still be substantially disrupted without introducing the most visible artifacts. Figure~\ref{fig:zero-shot-qual} further shows qualitative zero-shot classification examples, where clean images are classified correctly, whereas XSPA adversarial images induce clear misclassification under a thin X-shaped perturbation while preserving a largely natural visual appearance, consistent with Table~\ref{tab:zero-shot}.

\vspace{-0.35em}
\subsection{Grad-CAM-Based Attention Heatmap Analysis}

\begin{figure}[!t]
\centering
\includegraphics[width=\columnwidth]{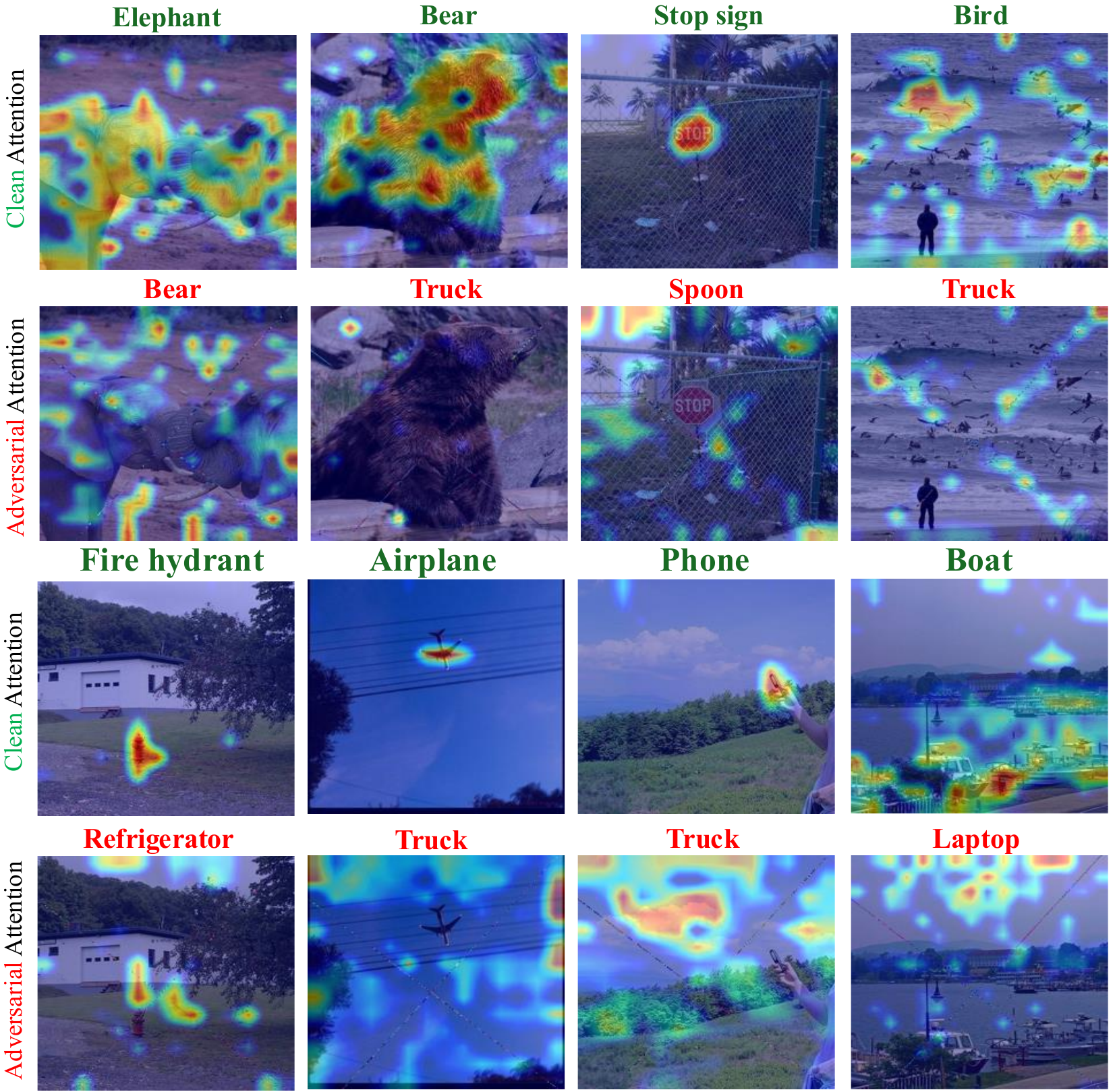}
\caption{Grad-CAM heatmaps for clean and adversarial samples. Adversarial responses become more diffuse or shift away from the main object.}
\label{fig:gradcam-heatmap}
\end{figure}

We use Grad-CAM to compare attention in clean and adversarial samples \cite{selvaraju2017_gradcam}. As shown in Figure~\ref{fig:gradcam-heatmap}, clean images concentrate responses on semantically decisive object regions, whereas XSPA produces more diffuse or displaced activation across both foreground-dominant and small-target scenes. This shift is most evident when clean maps form compact peaks, whereas adversarial maps fragment into several weaker responses distributed across less relevant spatial regions. The pattern spans large objects such as elephants and bears and localized targets such as stop signs, airplanes, phones, and boats, indicating that the attention shift is not confined to a particular object scale or scene complexity. Despite being confined to a fixed sparse X-shaped support, the perturbation redirects attention toward background or irrelevant structures, suggesting that its influence extends beyond the modified pixels to internal discriminative features. This redistribution is consistent with the performance degradation in our main experiments: XSPA disrupts the visual evidence used for recognition across varied object scales and scene complexity while largely preserving natural appearance. Together, these shifts consistently weaken object-centered spatial evidence across examples. The effect spans diverse scenes.

\noindent\textbf{Image Captioning.} Table~\ref{tab:caption} presents captioning results. Under the judge-based metric, XSPA yields the lowest consistency across six models, demonstrating disruption beyond classification in open-ended generation. Consistency falls to 46.40\% and 44.20\% on LLaVA-1.5 and LLaVA-1.6, and to 11.60\%, 21.90\%, 59.00\%, and 40.90\% on OpenFlamingo, BLIP-2 (FlanT5XL ViT-L), BLIP-2 (FlanT5XL), and InstructBLIP (FlanT5XL), respectively. OpenFlamingo shows the largest decrease, dropping 58.60 points from clean samples. The attack transfers across instruction-tuned and non-instruction-tuned systems: the strongest degradation appears on OpenFlamingo and BLIP-2 (FlanT5XL ViT-L), while LLaVA-1.5, LLaVA-1.6, and InstructBLIP still exhibit substantial drops. In many cases, generated captions no longer preserve object category, local attributes, or scene relations, even though the perturbation remains visually sparse and structurally constrained; this tendency is consistent across both short and descriptive captions. Overall, the results suggest that sparse geometric perturbations can propagate from visual features to sentence-level semantics and induce transferable caption drift across diverse model families.

\begin{table*}[h]
\caption{Image captioning results on six vision-language models under different adversarial attacks. Lower GPT-4 consistency indicates more effective attacks. Numbers in parentheses denote the absolute drop relative to clean samples. Bold indicates the best result.}
\label{tab:caption}
\centering
\resizebox{0.95\textwidth}{!}{%
\begin{tabular}{llcccccc}
\toprule
Image Encoder & Model & Params & Clean & NaturalLightAttack & ShadowAttack & ITA & Ours \\
\midrule
\multirow{2}{*}{OpenAI CLIP ViT-L/14} & LLaVA-1.5 & 7B & 78.60 & 77.00($\downarrow$1.60) & 74.60($\downarrow$4.00) & 63.73($\downarrow$14.87) & \textbf{46.40}($\downarrow$\textbf{32.20}) \\
\cmidrule(lr){2-8}
 & LLaVA-1.6 & 7B & 72.10 & 71.70($\downarrow$0.40) & 71.17($\downarrow$0.93) & 61.60($\downarrow$10.50) & \textbf{44.20}($\downarrow$\textbf{27.90}) \\
\midrule
\multirow{4}{*}{EVA-CLIP ViT-G/14} & OpenFlamingo & 3B & 70.20 & 69.53($\downarrow$0.67) & 67.80($\downarrow$2.40) & 53.93($\downarrow$16.27) & \textbf{11.60}($\downarrow$\textbf{58.60}) \\
\cmidrule(lr){2-8}
 & BLIP-2 (FlanT5XL ViT-L) & 3.4B & 75.10 & 70.77($\downarrow$4.33) & 68.57($\downarrow$6.53) & 60.93($\downarrow$14.17) & \textbf{21.90}($\downarrow$\textbf{53.20}) \\
\cmidrule(lr){2-8}
 & BLIP-2 (FlanT5XL) & 4.1B & 74.96 & 71.27($\downarrow$3.69) & 68.80($\downarrow$6.16) & 62.01($\downarrow$12.95) & \textbf{59.00}($\downarrow$\textbf{15.96}) \\
\cmidrule(lr){2-8}
 & InstructBLIP (FlanT5XL) & 4.1B & 76.50 & 72.07($\downarrow$4.43) & 69.77($\downarrow$6.73) & 63.20($\downarrow$13.30) & \textbf{40.90}($\downarrow$\textbf{35.60}) \\
\bottomrule
\end{tabular}%
}
\end{table*}

\begin{table*}[h]
\caption{VQA results on six vision-language models under different adversarial attacks. Lower GPT-4 correctness indicates more effective attacks. Numbers in parentheses denote the absolute drop relative to clean samples. Bold indicates the best result.}
\label{tab:vqa}
\centering
\resizebox{0.95\textwidth}{!}{%
\begin{tabular}{llcccccc}
\toprule
Image Encoder & Model & Params & Clean & NaturalLightAttack & ShadowAttack & ITA & Ours \\
\midrule
\multirow{2}{*}{OpenAI CLIP ViT-L/14} & LLaVA-1.5 & 7B & 68.00 & 68.00(-) & 67.00($\downarrow$1.00) & 48.00($\downarrow$20.00) & \textbf{47.45}($\downarrow$\textbf{20.55}) \\
\cmidrule(lr){2-8}
 & LLaVA-1.6 & 7B & 64.00 & 63.00($\downarrow$1.00) & 64.00(-) & \textbf{43.00}($\downarrow$\textbf{21.00}) & 48.87($\downarrow$15.13) \\
\midrule
\multirow{4}{*}{EVA-CLIP ViT-G/14} & OpenFlamingo & 3B & 45.00 & 39.00($\downarrow$6.00) & 44.00($\downarrow$1.00) & 19.00($\downarrow$26.00) & \textbf{14.29}($\downarrow$\textbf{30.71}) \\
\cmidrule(lr){2-8}
 & BLIP-2 (FlanT5XL ViT-L) & 3.4B & 63.00 & 58.00($\downarrow$5.00) & 50.00($\downarrow$13.00) & 38.00($\downarrow$25.00) & \textbf{18.75}($\downarrow$\textbf{44.25}) \\
\cmidrule(lr){2-8}
 & BLIP-2 (FlanT5XL) & 4.1B & 54.00 & 53.00($\downarrow$1.00) & 54.00(-) & 33.00($\downarrow$21.00) & \textbf{19.78}($\downarrow$\textbf{34.22}) \\
\cmidrule(lr){2-8}
 & InstructBLIP (FlanT5XL) & 4.1B & 68.00 & 64.00($\downarrow$4.00) & 62.00($\downarrow$6.00) & \textbf{44.00}($\downarrow$\textbf{24.00}) & 44.67($\downarrow$23.33) \\
\bottomrule
\end{tabular}%
}
\end{table*}

\FloatBarrier

\begingroup
\setlength{\stripsep}{6pt plus 1pt minus 1pt}
\begin{strip}
\centering
\captionsetup{type=figure,skip=3pt}
\includegraphics[width=0.97\textwidth]{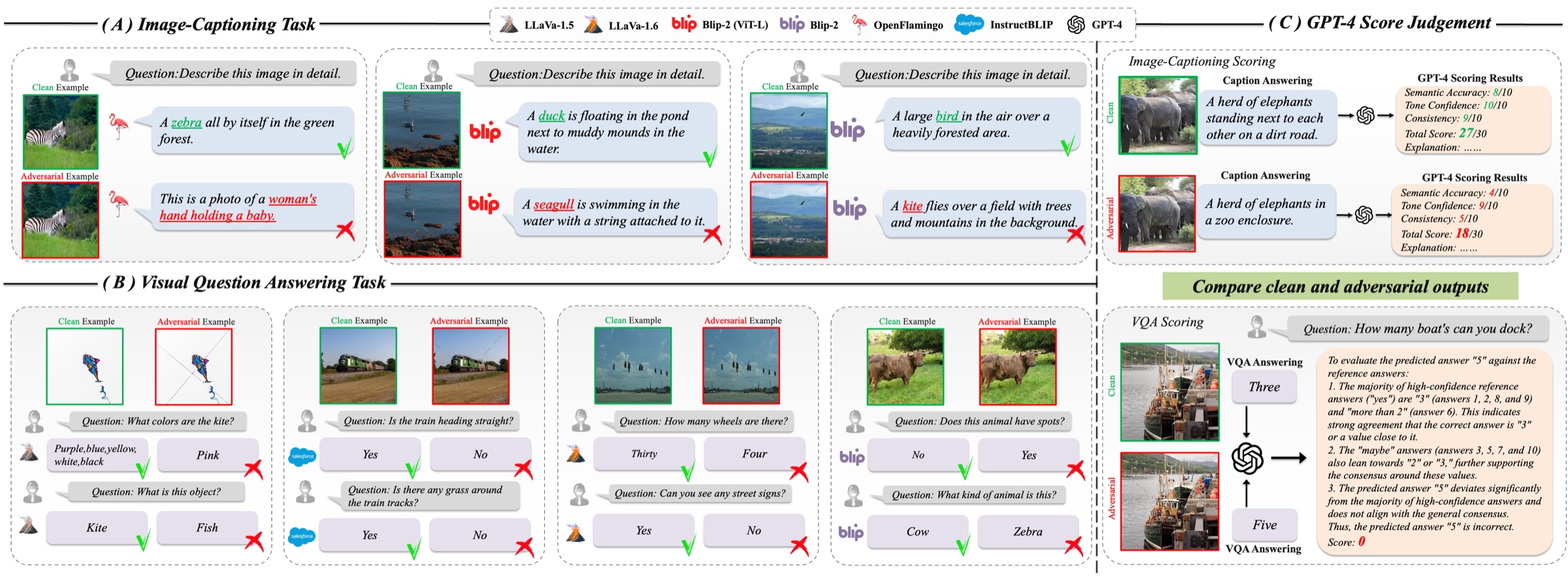}
\Description{Qualitative captioning and VQA examples comparing clean and adversarial inputs across multiple vision-language models. Adversarial samples induce semantic drift in captions and incorrect answers in VQA.}
\caption{Qualitative examples on image captioning and visual question answering. Compared with clean outputs, XSPA adversarial outputs exhibit clear semantic drift in captions and more error-prone answers in VQA, leading to lower GPT-4-based scores.}
\label{fig:vqa-caption-qual}
\end{strip}
\endgroup

\noindent\textbf{Visual Question Answering.} Table~\ref{tab:vqa} reports the VQA results. XSPA achieves the lowest correctness on four of the six models and remains competitive on the other two, indicating that its semantic objectives extend to question-conditioned generation. Specifically, correctness falls to 47.45\% and 48.87\% on LLaVA-1.5 and LLaVA-1.6, and to 14.29\%, 18.75\%, 19.78\%, and 44.67\% on OpenFlamingo, BLIP-2 (FlanT5XL ViT-L), BLIP-2 (FlanT5XL), and InstructBLIP (FlanT5XL), respectively. The largest drop, 44.25 percentage points, occurs on BLIP-2 (FlanT5XL ViT-L). Compared with NaturalLightAttack, ShadowAttack, and ITA, XSPA performs best on three of the four EVA-CLIP-based models, while ITA is slightly stronger on LLaVA-1.6 and InstructBLIP. The trend holds across dialogue-style and encoder-decoder VLMs, suggesting that XSPA disrupts visual evidence used for answer selection rather than merely degrading text fluency. Figure~\ref{fig:vqa-caption-qual} provides qualitative evidence for both tasks. Clean captions generally preserve the main object, scene, and attributes, whereas adversarial captions drift in object identity, material, or fine-grained attributes. Adversarial VQA outputs produce more category, color, count, and local-attribute errors across decoder families, including on simple questions. This indicates that semantic disruption is not confined to difficult or ambiguous prompts. The GPT-4-based judgments support transfer from the surrogate CLIP space to downstream generation and reasoning.

\begin{figure}[!t]
\centering
\captionsetup{skip=3pt}
\includegraphics[width=\columnwidth]{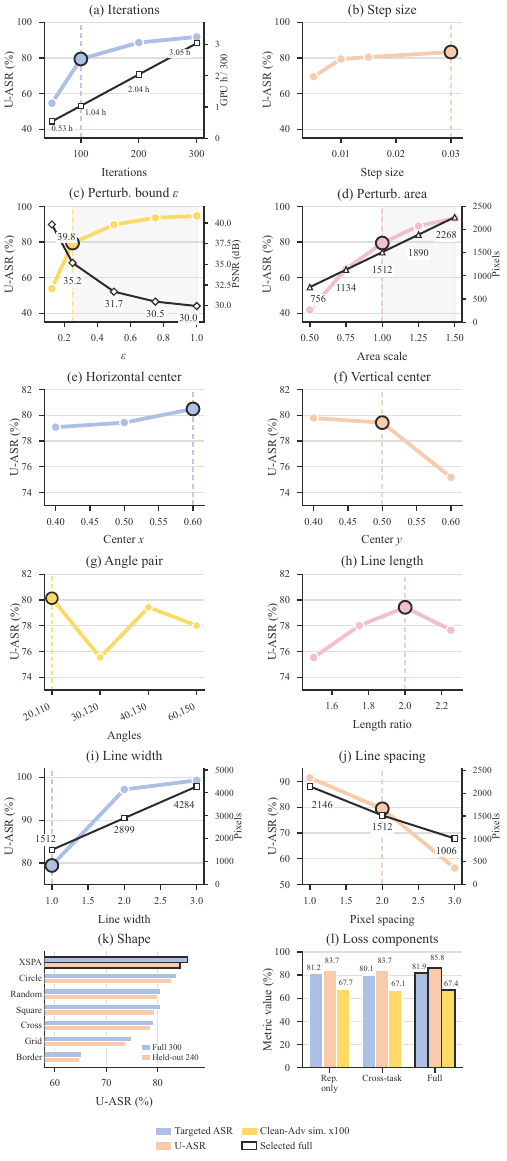}
\Description{Unified twelve-panel ablation figure in a single-column layout with two plot columns and six rows. Panels a to d vary optimization and perturbation settings; panels e to j vary X-support geometry; panel k compares sparse support shapes; panel l compares loss components.}
\caption{Unified single-column $2\times6$ ablations: (a)--(d) optimization and perturbation, (e)--(j) X geometry, (k) shape, and (l) loss components.}
\label{fig:ablation-unified}
\end{figure}

\subsection{Ablation Studies}

We study optimization and perturbation budgets, geometry and shape, loss design, and cross-dataset behavior. Unless varied, ablations use OpenCLIP ViT-B/16 on 300 images with 100 steps, a 0.03 step size, \(\epsilon=0.25\), and 1,532 pixels.

\noindent\textbf{Optimization, Geometry, Shape, and Loss Design.}
Figure~\ref{fig:ablation-unified} consolidates twelve ablations into a single-column grid with two plot columns and six rows. Panels (a)--(d) show that more iterations, a larger perturbation bound, or a larger area strengthen the attack, but increase GPU time, reduce PSNR, or modify more pixels. The saturation at larger iteration counts also indicates diminishing returns, supporting the selected operating point as a practical compromise between optimization cost, perturbation sparsity, and attack effectiveness for downstream evaluation. Panels (e)--(j) show moderate sensitivity to center, angle, and length, while larger line width or smaller pixel spacing sharply increases U-ASR together with support density. Center and angle sweeps remain flat, whereas width and spacing change sharply, identifying support density as the dominant geometric factor and exposing a trade-off between attack success and feature preservation. This separation helps distinguish robust design choices from budget-driven gains. Panel (k) compares seven shapes under the same 1,532-pixel budget at the final step size of 0.03; XSPA achieves the highest U-ASR on both Full 300 and Held-out 240. Panel (l) shows that no loss alternative dominates targeted ASR, U-ASR, and feature separation simultaneously, so the selected configuration balances attack strength, cost, and sparsity.

\noindent\textbf{Cross-Dataset and Multi-Encoder Behavior.}
Table~\ref{tab:ablation-cross-dataset} reports U-ASR on our original COCO-300 evaluation set and three larger evaluation sets across four CLIP-style encoders. Performance is highest on ImageNet-1K but lower on COCO and ImageNet-R, suggesting that dataset composition and encoder pretraining shape transferability rather than image count alone under the same attack protocol. XSPA remains effective beyond the original 300-image setting, while the variation across datasets and encoders makes the transfer boundary explicit rather than implying uniform vulnerability.

\begin{table}[!t]
\centering
\caption{Cross-dataset U-ASR (\%) across CLIP encoders. COCO-300 contains 300 images; all other datasets contain 1,000 images.}
\label{tab:ablation-cross-dataset}
\setlength{\tabcolsep}{2.2pt}
\resizebox{\columnwidth}{!}{%
\begin{tabular}{@{}lccccc@{}}
\toprule
\textbf{Dataset} &
\shortstack{\textbf{OpenAI}\\\textbf{L/14}} &
\shortstack{\textbf{OpenCLIP}\\\textbf{B/16}} &
    \shortstack{\textbf{Meta-CLIP}\\\textbf{L/14}} &
\shortstack{\textbf{EVA-CLIP}\\\textbf{ViT-G/14}} &
\textbf{Average} \\
\midrule
COCO-300 & 59.33 & 70.00 & 39.33 & 36.29 & \textbf{51.24} \\
COCO-1000 & 79.01 & 62.12 & 53.13 & 29.40 & \textbf{55.92} \\
ImageNet-1K & 99.58 & 97.55 & 98.19 & 84.73 & \textbf{95.01} \\
ImageNet-R & 73.56 & 66.89 & 54.34 & 56.48 & \textbf{62.82} \\
\bottomrule
\end{tabular}%
}
\end{table}

\FloatBarrier

\section{Conclusion}
\label{sec:conclusion}

This paper presents XSPA, a sparse X-shaped attack combining CLIP supervision, cross-task guidance, source-semantic suppression, and smoothness regularization. Across four CLIP encoders and multiple VLMs, XSPA degrades zero-shot classification, captioning, and VQA on COCO. Ablations examine optimization budgets, perturbation limits, geometry, and loss design. Results reveal VLM vulnerability to geometrically constrained semantic perturbations and motivate structure-aware defenses and robustness evaluation.

\bibliography{references_75_generated}

\clearpage
\appendix
\section{Additional Methodological Details}
\label{app:method-details}

\subsection{Mask Construction}

For the support \(\Omega_M\) defined in the main text, the binary mask is
\begin{equation}
M_{h,w}=\mathbf{1}\bigl[(h,w)\in\Omega_M\bigr].
\end{equation}
We construct \(\Omega_M\) by sampling the two diagonal centerline paths, rasterizing their coordinates, and applying a width-\(b\) neighborhood expansion around every sampled point. Clipping the expanded coordinates to the image boundary produces the final mask \(M\) and the ordered centerline path set \(\mathcal{P}\) used by the continuity regularizer.

\subsection{Complete Mask-constrained Optimization}

The current adversarial image and its diversified input are
\begin{equation}
\begin{aligned}
x^{adv,(t)}
&=\operatorname{clip}(x+M\odot\delta^{(t)},0,1),\\
\tilde{x}^{adv,(t)}
&=D(x^{adv,(t)}).
\end{aligned}
\end{equation}
Following DI-FGSM \cite{author2019_improving} and MI-FGSM \cite{author2018_boosting}, the gradient and momentum are
\begin{equation}
\begin{aligned}
g^{(t)}
&=\nabla_{\delta}\mathcal{L}\bigl(\tilde{x}^{adv,(t)}\bigr),\\
m^{(t+1)}
&=\mu m^{(t)}
+\frac{g^{(t)}}{\operatorname{mean}(|g^{(t)}|)+10^{-12}}.
\end{aligned}
\end{equation}
We then take a sign step and project it back to the feasible set:
\begin{equation}
\begin{aligned}
\tilde{\delta}^{(t+1)}
&=\delta^{(t)}-\alpha\,\operatorname{sign}\bigl(m^{(t+1)}\bigr),\\
\delta^{(t+1)}
&=M\odot\operatorname{clip}
\bigl(\tilde{\delta}^{(t+1)},-\epsilon,\epsilon\bigr).
\end{aligned}
\end{equation}

\begin{algorithm}[H]
\caption{Complete Pseudocode of XSPA}
\label{alg:xspa}
\footnotesize
\begin{algorithmic}[1]
\Require Image \(x\), mask \(M\), surrogate \(f\), text pool \(\mathcal{T}\), iterations \(N\), step size \(\alpha\), momentum \(\mu\), bound \(\epsilon\)
\Ensure Best adversarial sample \(x^{best}\)
\State Initialize \(\delta^{(0)}\gets0\), \(m^{(0)}\gets0\), and \(x^{best}\gets x\)
\State Build source and target text pools from \(\mathcal{T}\)
\For{\(t=0\) to \(N-1\)}
    \State \(x^{adv,(t)}\gets\operatorname{clip}(x+M\odot\delta^{(t)},0,1)\)
    \State Apply input diversity to \(x^{adv,(t)}\)
    \State Encode image and text features with \(f\)
    \State Compute \(\mathcal{L}_{clip}\), \(\mathcal{L}_{tar}\), \(\mathcal{L}_{src}\), and \(\mathcal{L}_{smooth}\)
    \State Form the joint objective \(\mathcal{L}\)
    \State \(g^{(t)}\gets\nabla_{\delta}\mathcal{L}\)
    \State \(m^{(t+1)}\gets\mu m^{(t)}+g^{(t)}/(\operatorname{mean}(|g^{(t)}|)+10^{-12})\)
    \State \(\tilde{\delta}^{(t+1)}\gets\delta^{(t)}-\alpha\,\operatorname{sign}(m^{(t+1)})\)
    \State Project \(\tilde{\delta}^{(t+1)}\) to obtain \(\delta^{(t+1)}\)
    \State Update \(x^{best}\)
\EndFor
\State \Return \(x^{best}\)
\end{algorithmic}
\end{algorithm}

\section{Evaluation and Reproducibility Details}
\label{app:evaluation-details}

The evaluation universe contains 300 COCO validation images spanning the COCO-80 label space. XSPA is optimized with white-box access to the surrogate encoder, while downstream captioning and VQA models are evaluated without adapting the perturbation to those models. Classification metrics use all 300 image IDs. The strict targeted VQA plan contains 380 questions on 224 of the 300 images; we state the smaller denominator whenever this plan is used.

The reviewer-oriented comparison freezes OpenCLIP ViT-B/16, the image order, optimization loss, and all attack hyperparameters. The only change between XSPA and the saliency Top-$k$ baseline is support selection: XSPA uses its fixed geometric support, whereas Top-$k$ selects the 1,532 clean-image pixels with the largest attack-loss gradient magnitude. Unless a quantity is itself varied, the configuration in Table~\ref{tab:appendix-defaults} is held fixed. This design prevents attack strength from being attributed to unequal pixel budgets or optimization effort.

\begin{table}[H]
\centering
\caption{Default configuration for the frozen ablation protocol.}
\label{tab:appendix-defaults}
\small
\begin{tabular}{@{}ll@{}}
\toprule
\textbf{Setting} & \textbf{Default value} \\
\midrule
Evaluation set & COCO-300 \\
Surrogate encoder & OpenCLIP ViT-B/16 \\
Input resolution & \(384\times384\) \\
Perturbation support & 1,532 pixels (approximately 1.04\%) \\
Optimization steps & 100 \\
Step size & 0.03 \\
\(L_\infty\) bound & 0.25 \\
\bottomrule
\end{tabular}
\end{table}

For zero-shot classification, primary attack success rate (ASR) is targeted success when a valid predefined target exists and untargeted success otherwise. All 300 images in the matched-budget study have valid targets, so primary and targeted ASR coincide. Untargeted ASR is reported separately. Statistical uncertainty is computed by resampling image IDs while retaining all optimization seeds within an image cluster; the 900 seed-image observations are not treated as independent samples.

\section{Matched-budget Strong Baseline and Seed Stability}
\label{app:matched-budget}

Table~\ref{tab:matched-budget-seeds} reports three independent optimization seeds under the frozen protocol. Saliency Top-$k$ achieves a primary ASR of $94.44\pm0.38$\%, compared with $83.44\pm0.77$\% for XSPA. The paired Top-$k$-minus-XSPA difference is 11.00 percentage points, with a 20,000-replicate image-cluster bootstrap 95\% confidence interval of $[7.78,14.44]$ points. For untargeted ASR, the corresponding difference is 8.00 points with a 95\% confidence interval of $[5.00,11.11]$. Exact paired McNemar tests are significant for every seed after Holm correction; the largest adjusted $p$-values are $8.70\times10^{-6}$ for primary ASR and $7.17\times10^{-4}$ for untargeted ASR.

\begin{table}[H]
\centering
\caption{Matched-budget ASR for saliency Top-$k$ and XSPA on the same 300 image IDs. Every run uses a 1,532-pixel support, $L_\infty=0.25$, 100 steps, and step size 0.03. Values are percentages.}
\label{tab:matched-budget-seeds}
\footnotesize
\setlength{\tabcolsep}{3pt}
\begin{tabular}{@{}llrrrr@{}}
\toprule
\textbf{Method} & \textbf{ASR} & \textbf{3407} & \textbf{3408} & \textbf{3409} & \textbf{Mean $\pm$ SD} \\
\midrule
Top-$k$ & Targeted & 94.67 & 94.00 & 94.67 & $94.44\pm0.38$ \\
XSPA & Targeted & 83.00 & 83.00 & 84.33 & $83.44\pm0.77$ \\
Top-$k$ & Untargeted & 95.00 & 94.67 & 94.67 & $94.78\pm0.19$ \\
XSPA & Untargeted & 86.67 & 86.67 & 87.00 & $86.78\pm0.19$ \\
\bottomrule
\end{tabular}
\end{table}

These results establish a narrow conclusion. The X-shaped support is effective under a severe sparsity constraint, but it is not the strongest support selector at the same pixel budget. Accordingly, the geometric contribution should be interpreted as a controlled structured-support stress test rather than an attack-strength advantage over adaptive saliency selection.

\section{Perceptual Distortion and Runtime}
\label{app:perceptual-runtime}

We compute full-resolution PSNR, SSIM, AlexNet-LPIPS, mean absolute error (MAE), changed-pixel ratio, and realized $L_\infty$ on all 300 paired image IDs for all three seeds. Table~\ref{tab:perceptual-runtime} reports method means obtained by first aggregating within each seed; uncertainty for the XSPA-minus-Top-$k$ difference is again obtained by image-cluster bootstrap. Runtime uses seeds 3408 and 3409 only because the frozen XSPA seed-3407 artifact predates per-image timing. The visual metrics retain all three seeds.

\begin{strip}
\centering
\captionof{table}{Perceptual distortion and runtime under the matched-budget protocol. Differences are XSPA minus saliency Top-$k$. Higher is better for PSNR and SSIM; lower is better for all other rows.}
\label{tab:perceptual-runtime}
\small
\begin{tabular}{@{}lrrrr@{}}
\toprule
\textbf{Metric} & \textbf{Top-$k$ mean $\pm$ SD} & \textbf{XSPA mean $\pm$ SD} & \textbf{Difference} & \textbf{95\% cluster CI} \\
\midrule
PSNR (dB) & $34.5981\pm0.0093$ & $34.1714\pm0.0113$ & $-0.4267$ & $[-0.4918,-0.3641]$ \\
SSIM & $0.96060\pm0.00007$ & $0.97285\pm0.00003$ & $+0.01225$ & $[+0.01132,+0.01321]$ \\
LPIPS-Alex & $0.08575\pm0.00016$ & $0.06516\pm0.00008$ & $-0.02059$ & $[-0.02359,-0.01767]$ \\
MAE & $0.001680\pm0.000003$ & $0.001798\pm0.000003$ & $+0.000118$ & $[+0.000100,+0.000137]$ \\
Changed pixels (\%) & $1.0353\pm0.0001$ & $1.0383\pm0.0000$ & $+0.0030$ & $[+0.0017,+0.0046]$ \\
Realized $L_\infty$ & $0.25098\pm0.00000$ & $0.25098\pm0.00000$ & $0.00000$ & $[0.00000,0.00000]$ \\
Attack time (s/image) & $12.455\pm0.191$ & $12.354\pm0.174$ & $-0.102$ & $[-0.116,-0.087]$ \\
\bottomrule
\end{tabular}
\end{strip}

The perceptual evidence is mixed rather than uniformly favorable. XSPA improves SSIM and LPIPS, indicating better structural and learned-feature similarity, whereas Top-$k$ has higher PSNR and lower MAE. Both methods realize the same $L_\infty$ bound and nearly the same support fraction, and their runtime difference is small relative to the approximately 12-second per-image cost. We therefore use ``sparse'' and ``low-coverage'' as directly supported descriptors and do not treat attack success as evidence of imperceptibility.

\section{Same-image Cross-task Joint Success}
\label{app:joint-success}

We additionally test whether classification, captioning, and VQA fail toward their predefined targets on the same adversarial image. This analysis uses seed 3407. Caption outputs are generated by BLIP-2 FlanT5-XL and VQA outputs by LLaVA-1.6 Mistral-7B. An image is jointly eligible when its clean caption does not already mention the target category and at least one planned VQA target answer is absent on the clean input. Of the 300 images, 224 have at least one strictly filtered targeted VQA question and 214 satisfy the joint clean-input eligibility rule.

\begin{table}[H]
\centering
\caption{Factorization of same-image targeted success on the 214 jointly eligible images. ``VQA any'' requires at least one eligible target answer; ``VQA all'' requires every eligible target answer for that image.}
\label{tab:joint-success}
\small
\begin{tabular}{@{}lrr@{}}
\toprule
\textbf{Component} & \textbf{XSPA} & \textbf{Saliency Top-$k$} \\
\midrule
CLIP targeted & 177/214 (82.71\%) & 203/214 (94.86\%) \\
Caption targeted & 0/214 (0.00\%) & 0/214 (0.00\%) \\
VQA any & 9/214 (4.21\%) & 8/214 (3.74\%) \\
VQA all & 3/214 (1.40\%) & 3/214 (1.40\%) \\
Joint with VQA any & 0/214 (0.00\%) & 0/214 (0.00\%) \\
Joint with VQA all & 0/214 (0.00\%) & 0/214 (0.00\%) \\
\bottomrule
\end{tabular}
\end{table}

Neither method achieves a three-task targeted success on this common denominator because the downstream caption model never introduces the predefined target category. The result does not support a claim of correlated targeted failure across all three downstream tasks. It instead separates strong surrogate classification effects from substantially weaker output-space transfer.

\section{Judge Calibration and Claim Boundary}
\label{app:judge-calibration}

The GPT-based evaluation is calibrated on an overlapping LLaVA-1.5 subset containing 30 clean and 30 XSPA outputs per task. Each item is independently scored by GPT-4 and GPT-4o-mini under the same task rubric. The bootstrap resamples item IDs and retains the clean/XSPA condition pair. This is an inter-GPT calibration study; it is not human-GPT agreement.

\begin{table}[H]
\centering
\caption{Inter-GPT judge calibration on 60 condition-items per task.}
\label{tab:judge-calibration}
\small
\begin{tabular}{@{}llr@{}}
\toprule
\textbf{Task} & \textbf{Reliability measure} & \textbf{Estimate [95\% CI]} \\
\midrule
Caption & Pearson $r$ & $0.580\;[0.255,0.801]$ \\
Caption & Spearman $\rho$ & $0.579\;[0.268,0.796]$ \\
Caption & ICC(A,1) & $0.301\;[0.120,0.443]$ \\
Caption & MAE (0--30 scale) & $3.683\;[3.050,4.367]$ \\
VQA & Exact agreement & $0.933\;[0.850,1.000]$ \\
VQA & Cohen's $\kappa$ & $0.815\;[0.526,1.000]$ \\
\bottomrule
\end{tabular}
\end{table}

VQA scoring shows high observed agreement, whereas caption scoring has only moderate rank correlation and low absolute-agreement ICC. GPT-4 assigns caption scores that are, on average, 3.283 points higher on the 0--30 scale than GPT-4o-mini. Because independent human ratings are not yet available, these data do not constitute human validation. We consequently treat judge-based caption differences as secondary evidence and avoid using them alone to support cross-task transfer claims.

\AppendixColumnBreak
\section{Statistical and Artifact Audit}
\label{app:artifact-audit}

All matched-budget summaries are reconstructed from per-image JSON records rather than terminal logs. The audit checks 300 unique IDs in every method-seed cell, verifies a 1,532-pixel mask on every image, and requires the method-specific mask variant to agree with the recorded support selector. Paired confidence intervals use 20,000 image-cluster bootstrap replicates with seed 20260731. Binary method comparisons use exact McNemar tests within each optimization seed and Holm correction across the three seeds for a metric. These choices preserve the paired image structure and prevent optimization replicates from inflating the effective sample size.

Overall, the evidence supports sparse surrogate attacks and a perceptual tradeoff, but not universal superiority, uniform imperceptibility, or reliable three-task targeted transfer. Human perceptual and human-GPT studies are incomplete and are not reported as findings.

\end{document}